\documentclass[11pt,a4paper]{article}

\usepackage[dvipsnames]{xcolor}
\usepackage[hyperref]{naaclhlt2019}

\usepackage{times}
\usepackage{latexsym}
\usepackage{url}

\usepackage{amssymb}
\usepackage{pifont}
\usepackage{amssymb}
\usepackage{amsmath}

\usepackage{cleveref}
\usepackage{stmaryrd}
\usepackage{graphicx}
\usepackage{adjustbox}
\usepackage{listings}
\usepackage{caption,setspace}
\usepackage{diagbox} 
\usepackage{booktabs} 
\usepackage{rotating}
\usepackage{array,multirow}
\usepackage{hhline}
\usepackage{todonotes}
\usepackage{flushend}
\usepackage[inline]{enumitem}
\usepackage{csquotes}
\usepackage{subcaption}
\usepackage[shortcuts]{extdash} 
\usepackage[normalem]{ulem} 

\newcommand{\etc}{etc.\ }
\newcommand{\eg}{e.g., }
\newcommand{\ie}{i.e., }
\newcommand{\figref}[1]{Fig.~\ref{#1}}    
\newcommand{\Figref}[1]{Figure~\ref{#1}}  
\newcommand{\tabref}[1]{Table~\ref{#1}}

\newcommand{\secref}[1]{Section~\ref{#1}}

\newcommand{\cmark}{\ding{51}}%
\newcommand{\xmark}{\ding{55}}%

\newcommand{\Fone}{{F$_{1}$}}

\makeatletter
\def\Url@twoslashes{\mathchar`\/\@ifnextchar/{\kern-.2em}{}}
\g@addto@macro\UrlSpecials{\do\/{\Url@twoslashes}}
\makeatother

\newif\iftrackchanges

\iftrackchanges
\usepackage{color}
\newcommand{\revone}[1]{\textcolor{RoyalBlue}{[R1]~\textbf{#1}}}

\newcommand{\revthree}[1]{\textcolor{RoyalBlue}{[R3]~\textbf{#1}}}
\newcommand{\oldtext}[1]{\textcolor{RedViolet}{\sout{#1}}}
\newcommand{\newtext}[1]{\textcolor{RoyalBlue}{\textbf{#1}}}
\else 
\newcommand{\revone}[1]{#1}

\newcommand{\revthree}[1]{#1}
\newcommand{\oldtext}[1]{}
\newcommand{\newtext}[1]{#1}
\fi

\aclfinalcopy 
\def\aclpaperid{367} 

\title{Sub\-/event detection from Twitter streams as\\ a sequence labeling problem}

\author{Giannis Bekoulis\qquad Johannes Deleu\qquad Thomas Demeester\qquad Chris Develder\\Ghent University -– imec, IDLab\\Department of Information Technology\\ \tt{firstname.lastname@ugent.be}}

\date{}

\begin{document}
\maketitle
\begin{abstract}


This paper introduces improved methods for sub\-/event detection in social media streams, by applying neural sequence models not only on the level of individual posts, but also directly on the stream level.
Current approaches to identify sub\-/events within a given event, \oldtext{(e.g.,}\newtext{such as} a goal during a soccer match\oldtext{)},
essentially do not exploit the sequential nature of social media streams.
We address this shortcoming by framing the sub\-/event detection problem in social media streams as a sequence labeling task and adopt a neural sequence architecture that explicitly accounts for the chronological order of posts.
Specifically, we
\begin{enumerate*}[label=(\roman*)]
\item establish a neural baseline that outperforms a graph-based state-of-the-art method for binary sub\-/event detection (2.7\% \newtext{micro-}{\Fone} improvement), as well as 
\item demonstrate superiority of a recurrent neural network model on the posts sequence level for labeled sub\-/events (2.4\% \newtext{bin-level} {\Fone} improvement over non-sequential models).
\end{enumerate*}
\end{abstract}

\section{Introduction}

Social media allow users to communicate via real-time postings and interactions, 
with Twitter as a notable example.
Twitter user posts\newtext{, \ie} \oldtext{(aka} tweets\oldtext{)}\newtext{,} are often related to events. These can be social events (concerts, research conferences, sports events, etc.), emergency situations (\eg terrorist attacks)~\cite{castillo:16}, etc. 
For a single event, multiple tweets are posted, by people with various personalities and social behavior.
Hence, even more so than (typically more neutral) traditional media, this 
implies many different perspectives, offering an interesting aggregated description.

Given this continuous and large stream of (likely duplicated) information in Twitter streams, and their noisy nature, it is challenging to keep track of the main parts of an event\newtext{, such as}\oldtext{(\eg} a soccer match\oldtext{)}.
Automating such extraction of different sub\-/events within an evolving event is known as sub\-/event detection~\cite{nichols:12}.
For tracking each of the sub\-/events, the timing aspect is an important concept (\ie consecutive tweets in time). 
\revone{Thus, a sequential model could successfully exploit chronological relations between the tweets in a Twitter stream as an informative feature for sub\-/event detection.} 

Several methods have been proposed for sub\-/event detection: clustering methods~\cite{pohl:12}, graph-based approaches~\cite{meladianos:15}, topic models~\cite{xing:16} and neural network architectures~\cite{wang:17}.
None of these studies exploits the chronological relation between consecutive tweets.
In contrast, our work does take into account that chronological order and we predict the presence and the type of a sub\-/event exploiting information from previous tweets. Specifically, we
\begin{enumerate*}[label=(\roman*)]
\item propose a \newtext{new neural} baseline model that outperforms the state-of-the-art performance on the
\oldtext{sports stream sub\-/event (presence/absence) detection problem,}
\newtext{binary classification problem of detecting the presence/absence of sub\-/events in a sports stream, }
\item establish a \newtext{new} reasonable baseline for predicting also the sub\-/event \emph{types},
\item explicitly take into account chronological information\newtext{,} \oldtext{(}\newtext{\ie the} relation among consecutive tweets\oldtext{)}\newtext{,} by framing sub\-/event detection as a sequence labeling problem on top of our baseline model, and
\item perform an \oldtext{extensive} experimental study, indicating the benefit of sequence labeling for sub\-/event detection in sports Twitter streams.
\end{enumerate*}

\begin{figure*}
\includegraphics[width=\textwidth]{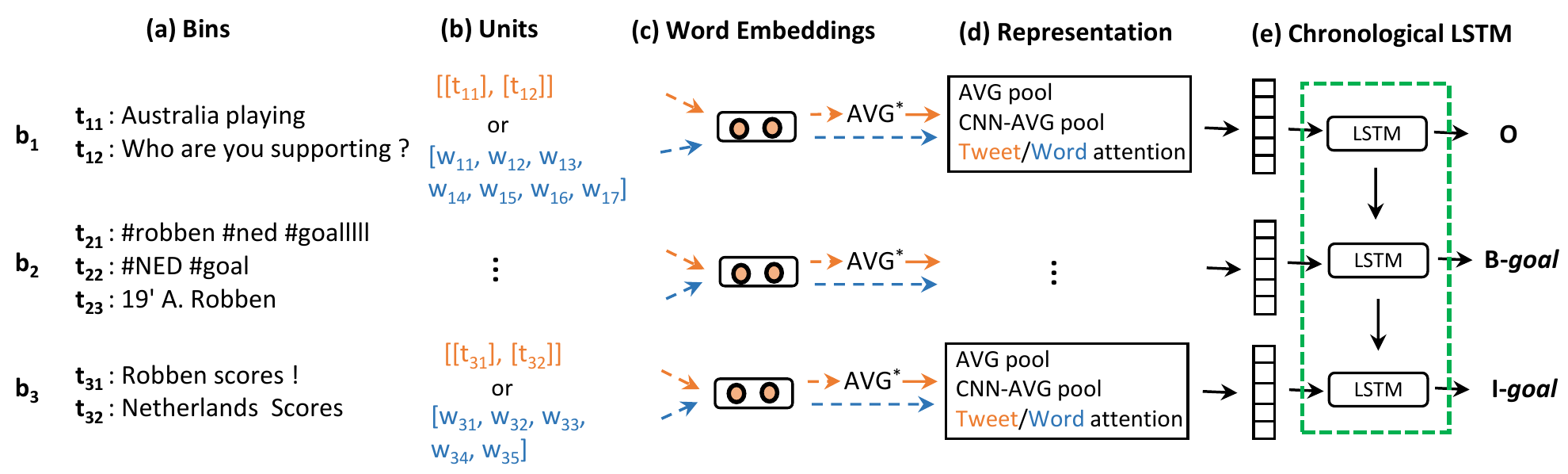}
\caption{Our sub\-/event detection model comprises:
\begin{enumerate*}[label=(\alph*)]
\item a bin layer,
\item a unit layer,
\item a word embeddings layer,
\item a representation layer and 
\item a \textcolor{ForestGreen}{\textbf{chronological LSTM}} layer to model the natural flow of the sub\-/events within the event.
\end{enumerate*}
We represent each bin using either
\begin{enumerate*}[label=(\roman*)]
\item a \textcolor{BurntOrange}{\textbf{tweet}}- or
\item a \textcolor{MidnightBlue}{\textbf{word}}-level
\end{enumerate*} 
representation.
The AVG$^*$ represents an average pool operation, performed either directly on the embeddings or on the tweet's LSTM representation.}
\label{fig:model}
\end{figure*}

\section{Related work}
\revone{Twitter streams have been extensively studied in various contexts, such as sentiment analysis~\cite{kouloumpis:11}, stock market prediction~\cite{nguyen:15} and traffic detection~\cite{andrea:15}.
Specifically,} for sub\-/event detection in Twitter, several approaches have been tried. 
\emph{Unsupervised methods} \newtext{such as}\oldtext{like} clustering aim to group similar tweets to detect specific sub\-/events~\cite{pohl:12,abhik:13} and use simple representations such as tf-idf weighting combined with a similarity measure.
Other unsupervised algorithms use topic modeling approaches, based on assumptions about the tweets' generation process~\cite{xing:16,srijith:17}.
Several methods~\cite{zhao:11,zubiaga:12,nichols:12} assume that a sub\-/event happens when there is a `burst', \newtext{\ie} a sudden  increase in the rate of tweets on the considered event, with \oldtext{a lot of}\newtext{many} people commenting \oldtext{about}\newtext{on} it. 
Recently, neural network methods have used more complicated representations~\cite{wang:17,chen:18}. 
\revone{Also \textit{supervised methods} have been applied~\cite{sakaki:10,meladianos:18} for the sub\-/event detection task. These methods usually exploit graph-based structures or tf-idf weighting schemes}.
%
We believe to be the first to 
\begin{enumerate*}[label=(\roman*)]
\item exploit the chronological order of the Twitter stream and take into account its sequential nature, and 
\item frame the sub\-/event detection problem as a sequence labeling task.
\end{enumerate*}

\section{Model}

\subsection{Task definition}
The goal is, given a main event (\ie soccer match), to identify its core sub\-/events (\eg goals, kick-off, yellow cards) from Twitter streams. Specifically, we 
consider a \emph{supervised} setting, 
relying on
annotated data~\cite{meladianos:18}.

\subsection{Word- vs tweet-level representations}

Similar to previous works, we split a data stream into time periods~\cite{meladianos:18}: we form bins of tweets posted during consecutive time intervals.
E.g., for a soccer game, one-minute intervals (bins) lead to \revone{more than 90 bins, depending on the content before and after the game, halftime, stoppage time, and possibly some pre-game and post-game buffer}. Thus, for each bin, we predict either the presence/absence of a sub\-/event~(\secref{sec:baseline}) or the most probable sub\-/event type~(\secref{sec:sequence_labeling}), depending on the evaluation scenario\oldtext{ (to compare to prior art)}. 

We consider representing the content of each bin either on
\begin{enumerate*}[label=(\roman*)]
\item a word-level or
\item a tweet-level
\end{enumerate*}
(see \figref{fig:model}). 
Formally, we assume that we have a set of $n$ bins $b_1,...,b_n$, where each bin $b_i$ consists of $m_i$ tweets and $k_i$ words (\ie all words of tweets in bin $b_i$).
Then, the \emph{tweet-level representation} of bin $b_i$ is symbolized as $t_{i1},..., t_{im_i}$, where $t_{im_i}$ is the $m_i^{\text{th}}$ tweet of bin $b_i$.
In the \emph{word-level representation}, we chronologically concatenate the words from the tweets in the bin: $w_{i1},..,w_{ik_i}$, where $w_{ik_i}$ is the $k_i^{\text{th}}$ word of bin $b_i$.

\subsection{Binary classification baseline}
\label{sec:baseline}
To compare with previous work~\cite{meladianos:18}, we establish a simple baseline for binary classification: presence/absence of a sub\-/event. For this case, we use as input the word-level representation of each bin. To do so, we use word embeddings (randomly initialized) with average (AVG) pooling~\cite{iyyer:15} in combination with a multilayer perceptron (MLP) for binary classification\oldtext{ (}\newtext{, }\ie presence/absence of a sub\-/event\oldtext{)}. Note that we experimented with pre-trained embeddings \newtext{as well as max-pooling}, but \newtext{those early experiments led to performance decrease compared to the presented baseline model.}\oldtext{ the performance decreased in our early experiments. We also experimented with max-pooling.}  
\oldtext{However, it appears}\newtext{We found} that training based on average bin representations works substantially better than with max-pooling, and we hypothesize that this is related to the noisy nature of the Twitter stream.


\subsection{Sequence labeling approach}
\label{sec:sequence_labeling}
Building on the baseline above, we establish a new architecture that is able to capture the sub\-/event types as well as their duration. 
We phrase sub\-/event detection in Twitter streams as a sequence labeling problem. This means we assume that the label of a bin is not independent of neighboring bin labels, given the chronological order of bins of the Twitter stream\oldtext{ (}\newtext{, }as opposed to independent prediction for each bin\oldtext{)}\newtext{ in the binary classification baseline above}. 
For instance, when a \emph{goal} is predicted as a label for bin $b_i$, then it is probable that the label of the next bin $b_{i+1}$ will also be \emph{goal}. \revone{Although a sub\-/event may occur instantly, an identified sub\-/event in a Twitter stream can span consecutive bins, \ie minutes: users may continue tweeting on a particular sub\-/event for relatively long time intervals.} For this reason, we apply the well-known BIO tagging scheme~\cite{ramshaw:95} for the sub\-/event detection problem. For example, the beginning of a \emph{goal} sub\-/event is defined as B-\emph{goal}, while I-\emph{goal} (inside) is assigned to every consecutive bin within the same sub\-/event, and the O tag (outside) to \oldtext{each}\newtext{every} bin \newtext{that is} not part of any sub\-/event. To propagate chronological information among bins, we \oldtext{employ}\newtext{adopt} an LSTM on the sequence of bins as illustrated in~\figref{fig:model}\oldtext{ (see }\newtext{, }layer (e)\oldtext{)}. 
\newtext{Note that this tagging approach assumes that sub\-/events do not overlap in time, \ie only at most one is ongoing in the Twitter stream at any point in time.}

\section{Experimental setup}
\label{sec:experimental_setup}
We evaluated our system\footnote{\url{https://github.com/bekou/subevent_sequence_labeling}} on \newtext{the dataset from \newcite{meladianos:18}, with tweets on }20 soccer matches from the 2010 and 2014 FIFA World Cups\newtext{, totalling over 2M pre-processed tweets filtered from 6.1M collected ones, comprising 185 events}.
The dataset includes a set of sub\-/events, such as \emph{goal}, \emph{kick-off}, \emph{half-time}, \etc  
To compare our \newtext{binary classification} \emph{baseline system} to previous methods~(\tabref{tab:results_baseline}), we use the same train/test splits as~\newcite{meladianos:18}, where 3 matches are used for training and 17 matches as test set. In this setting, we predict only the presence/absence of a sub\-/event. Similar to previous work, we count a sub\-/event as correct if at least one of its comprising bins has been classified as a sub\-/event.
For the experimental study of our proposed \emph{sequence labeling approach} for sub\-/event detection, where sub\-/event types are predicted, we have randomly split the test set into test (10 matches) and development (7 matches) sets. We use the development set to optimize the {\Fone} score for tuning of \oldtext{representation sizes}\revone{the model parameters, \ie the word/tweet embedding representation size, LSTM hidden state size, dropout probability}. We adopt 2 evaluation strategies. The first one, referred to as \emph{relaxed} evaluation, is commonly used in entity classification tasks~\cite{adel:17,bekoulis:18c,bekoulis:18b} and similar to the \newtext{binary classification} baseline system evaluation: score a multi-bin sub\-/event as correct if at least one of its comprising bin types (\eg \emph{goal}) is correct, assuming that the boundaries are given. The second evaluation strategy, \emph{bin-level}, is stricter: we count each bin individually, and check whether its sub\-/event type has been predicted correctly, similar to the token-based evaluation followed in~\newcite{bekoulis:18a}.

\section{Results}

\subsection{Baseline results}
\tabref{tab:results_baseline} shows the experimental results of our baseline model. The Burst baseline system is based on the tweeting rate in a specific time window (\ie bin) and if a threshold is exceed, the system identifies that a sub\-/event has occurred. We report evaluation scores as presented in~\newcite{meladianos:18}. The second approach is the graph-based method of~\newcite{meladianos:18}. We observe that our baseline system~(\secref{sec:baseline}) has a 1.2\% improvement in terms of macro-{\Fone}
and 2.7\% improvement in terms of micro-\Fone, compared to the graph-based model from \newcite{meladianos:18}, mainly due to increased precision, and despite the recall loss.

\begin{table}[b]
\centering
\resizebox{\columnwidth}{!}{%
\begin{tabular}{@{\extracolsep{4pt}}cccccccc@{}} 
 \toprule
 & \multicolumn{1}{c}{} & \multicolumn{3}{c}{Macro}  & \multicolumn{3}{c}{Micro}  \\
\cline{3-5}
\cline{6-8}
 & \multicolumn{1}{c}{Settings} & \multicolumn{1}{c}{P} & \multicolumn{1}{c}{R}& \multicolumn{1}{c}{\Fone} & \multicolumn{1}{c}{P}& \multicolumn{1}{c}{R} & \multicolumn{1}{c}{F$_1$}  \\
 \midrule
&Burst   & 78.00 &54.00& 64.00 &72.00& 54.00& 62.00 \\
&\newcite{meladianos:18}& 76.00& 75.00& 75.00& 73.00& 74.00& 73.00   \\
&  \textbf{Our binary classif.\ baseline}   &89.70 &69.99&\textbf{76.16} &83.65&69.05&\textbf{75.65} \\
\bottomrule
\end{tabular}
}
\caption{Comparing our neural network \newtext{binary classification} baseline \oldtext{method} \newtext{model} to state-of-the-art (P = precision, R = recall).}
\label{tab:results_baseline}
\end{table}
 
\subsection{Sequence labeling results}
\tabref{tab:results_duration} illustrates the predictive performance of our proposed model (\ie using the chronological LSTM) compared to models making independent predictions per bin. The upper part of~\tabref{tab:results_duration} contains models without the chronological LSTM. 
Our experiments study both \emph{word-level} and \emph{tweet-level} bin representations (see~\figref{fig:model}), as reflected in the `Word' vs.\ `Tweet' prefix, respectively, in the Model column of \tabref{tab:results_duration}.

The simplest \emph{word-level} representation uses the tf-idf weighting scheme (as in~\newcite{pohl:12}) followed by an MLP classifier.
For the other word-level models, we exploit several architectures: AVG pooling~\cite{iyyer:15}, a CNN followed by AVG pooling~\cite{kim:14} and hierarchical word-level attention~\cite{yang:16}.

\begin{table}
\centering
\resizebox{\columnwidth}{!}{%
\begin{tabular}{@{\extracolsep{4pt}}cccccccccc@{}} 
 \toprule
 & \multicolumn{1}{c}{} & \multicolumn{4}{c}{Bin-level}& \multicolumn{4}{c}{Relaxed}  \\
\cline{3-6}
\cline{7-10}
 & \multicolumn{1}{c}{Model} & \multicolumn{1}{c}{TL} & \multicolumn{1}{c}{P} & \multicolumn{1}{c}{R}& \multicolumn{1}{c}{F$_1$} & \multicolumn{1}{c}{TL} & \multicolumn{1}{c}{P}& \multicolumn{1}{c}{R} & \multicolumn{1}{c}{F$_1$}  \\
 \midrule
\parbox[c]{5mm}{\multirow{7}{*}{\rotatebox[origin=c]{90}{\parbox{2.6cm}{\centering without chronol. LSTM}}}} 
&Word-tf-idf &-   & 49.40 &52.06& 50.69& - &56.10& 56.10& 56.10 \\
&Word-AVG &-  & 51.40 &45.96& 48.53&- &56.10& 56.10& 56.10  \\
& Word-CNN-AVG &-  &56.93 &56.01& 56.47 &- &75.60& 75.60& 75.60 \\
& Word-attention &-  & 52.92 &58.71& 55.66 &- &86.59& 86.59& \textbf{86.59} \\
& Tweet-AVG &\cmark  & 49.04 &45.96& 47.45 &\cmark &62.19& 62.19& 62.19 \\ 
& Tweet-attention &\cmark  & 51.99 &42.37& 46.68 &\xmark &80.48& 80.48& 80.48 \\
& Tweet-CNN &\xmark  & 58.88 &51.17& 54.75&\xmark &70.73& 70.73& 70.73 \\
\midrule
\midrule
\parbox[c]{5mm}{\multirow{6}{*}{\rotatebox[origin=c]{90}{\parbox{2.6cm}{\centering with chronol. LSTM}}}}
&Word-AVG &- & 58.14 &58.35& 58.24 &- &71.95& 71.95& 71.95   \\
& Word-CNN-AVG  &-  & 60.89 &56.19& 58.45&- &60.97& 60.97& 60.97 \\ 
& Word-attention &-  & 52.99 &42.90 & 47.42 &- &60.97& 60.97& 60.97\\
& Tweet-AVG  &\xmark & 57.43 &60.32& \textbf{58.84} &\xmark &64.63& 64.63& 64.63 \\
& Tweet-attention &\cmark  &48.26 &52.24& 50.17 &\xmark &67.07&67.07& 67.07 \\ 
& Tweet-CNN 
&\xmark  &65.33 &49.73& 56.47 &\xmark &60.97& 60.97& 60.97 \\ 
\bottomrule
\end{tabular}
}
\caption{Comparison of our baseline methods in terms of micro \emph{bin-level} and \emph{relaxed} {\Fone} score with and without chronological LSTM (see \figref{fig:model}). 
The \cmark and \xmark~indicate whether the model uses a tweet-level LSTM (TL).}
\label{tab:results_duration}
\end{table}

For \emph{tweet-level} representations, we adopt similar architectures, where the AVG, CNNs and attention are performed on sentence level rather than on the word-level representation of the bin. In this scenario, we have also exploited the usage of sequential LSTMs to represent the tweets. 
When comparing models with and without tweet-level LSTMs, we report the strategy that yields the best results, indicated by {\cmark} and {\xmark} in the tweet-level LSTM (TL) columns of~\tabref{tab:results_duration}. 
We do not present results for applying sequential LSTMs on the word-level bin representation, because of slow training on the long word sequences.

\noindent\textbf{Benefit of chronological LSTM:} The bottom part of~\tabref{tab:results_duration} presents the results of the same models followed by a chronological LSTM to capture the natural flow of the stream as illustrated in~\figref{fig:model}. We report results as described in~\secref{sec:experimental_setup}, using the micro {\Fone} score with the two evaluation strategies (\emph{bin-level} and \emph{relaxed}). 
We observe that when using the chronological LSTM, the performance in terms of \emph{bin-level} {\Fone} score is substantially improved for almost every model. Note that the best model using the chronological LSTM (Tweet-AVG) achieves 2.4\% better {\Fone} than the best performing model without the use of chronological LSTM (Word-CNN-AVG). In most cases there is also a consistent improvement for both the precision and the recall metrics, which is thanks to the sequential nature of the upper level LSTM capturing the flow of the text.

\noindent\textbf{Limitations of \emph{relaxed} evaluation:} On the other hand, using the \emph{relaxed} evaluation strategy, we observe that
the best models are those without the chronological LSTM layer.
Yet, we consider the \emph{relaxed} evaluation strategy flawed for our scenario, despite the fact that
it has been used for entity classification tasks~\cite{bekoulis:18c,adel:17}. Indeed, it is not able to properly capture sub\-/events which are characterized by duration:
\eg if a model assigns a different label to each of the bins that together constitute a single sub\-/event,
then this sub\-/event counts as a true positive based on the \emph{relaxed} evaluation strategy (similar to the evaluation proposed by~\newcite{meladianos:18} and followed in~\tabref{tab:results_baseline}). Thus, in this work, we propose to use the~\emph{bin-level} evaluation\newtext{,} since it is a more natural way to measure the duration of a sub\-/event in a supervised sequence labeling \oldtext{problem} \newtext{setting}.
\oldtext{Note that the standard sequence labeling evaluation (types and boundaries correct) is not applicable in sub\-/event detection since the exact boundaries cannot be accurately predicted (as in the token-based sequence labeling), due to the noisy nature of Twitter streams.}
\newtext{Note that due to the noisy nature of Twitter streams, a tweet sequence spanning a particular sub\-/event is likely to contain also tweets that are not related to the given sub\-/event: a given bin inside the event may contain only a minority of tweets discussing the event. Therefore, we consider the standard sequence labeling evaluation (requiring to have types as well as boundaries correct) to be not applicable in sub\-/event detection.}

\begin{figure}
\includegraphics[width=\columnwidth]{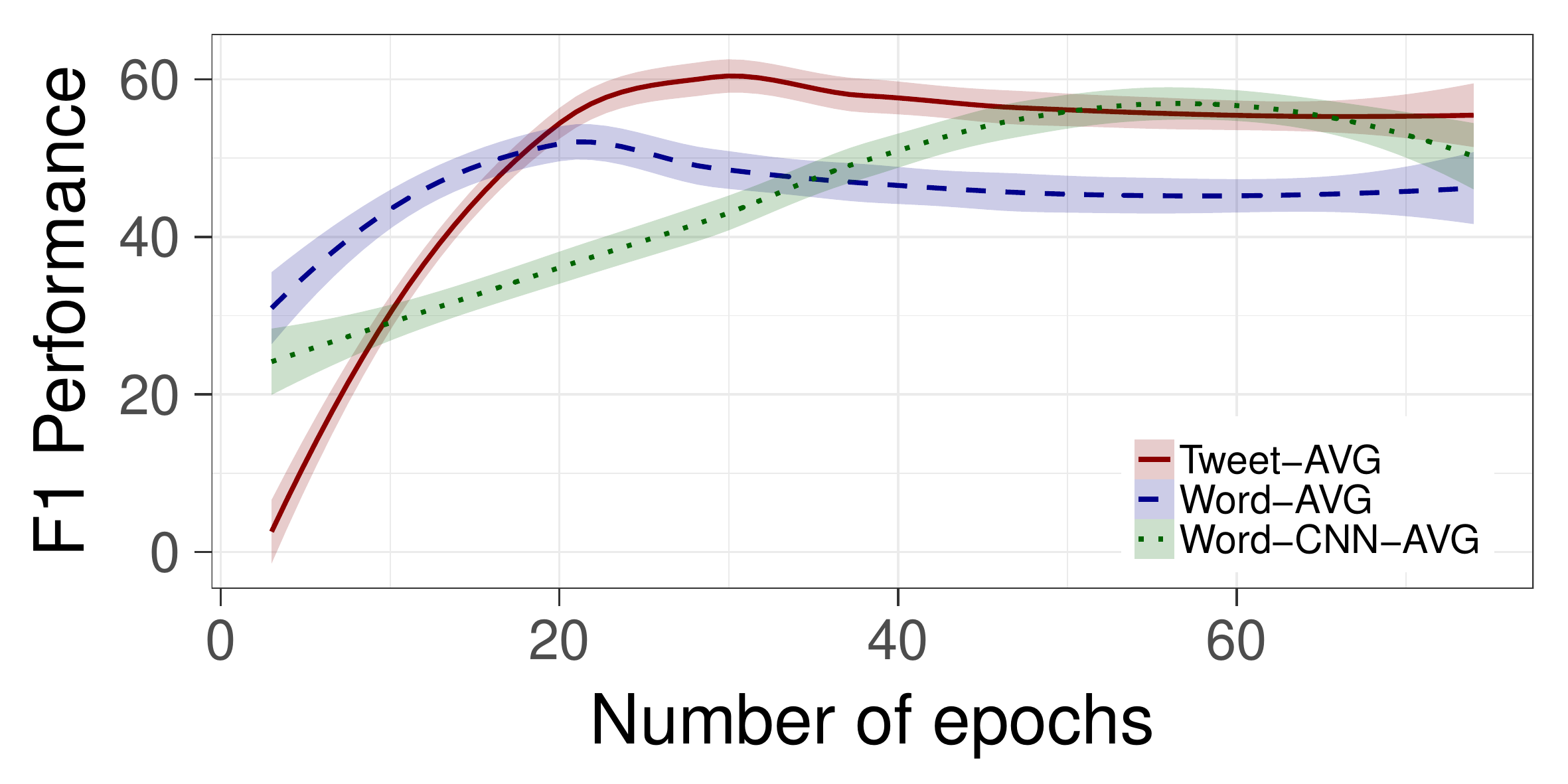}
\caption{\emph{Bin-level} {\Fone} performance of the three best performing models on the validation set with respect to the number of epochs. The smoothed lines (obtained by LOWESS smoothing) model the 
trends and the 95\% confidence intervals.}
\label{fig:dev_plot}
\end{figure}

\noindent\textbf{Performance comparison of the top-3 models:} \Figref{fig:dev_plot} shows the performance of our three best performing models in terms of \emph{bin-level} {\Fone} score on the validation set. The best performing model is the Tweet-AVG model since it attains its maximum performance even from the first training epochs. The Word-AVG model performs well from the first epochs, showing similar behavior to the Tweet-AVG model. This can be explained by the similar nature of the two models. The word-level CNN model attains maximum performance compared to the other two models in later epochs. Overall, we propose the use of the chronological LSTM with the Tweet-AVG model since this model does not rely on complex architectures and it gives consistent results.

\section{Conclusion}
In this work, we frame the problem of sub\-/event detection in Twitter streams as a sequence labeling task. Specifically, we 
\begin{enumerate*}[label=(\roman*)]
\item propose a binary classification baseline model that outperforms state-of-the-art approaches for sub\-/event detection (presence/absence), 
\item establish a strong baseline that additionally predicts sub\-/event \emph{types}, and then
\item extend this baseline model with the idea of exchanging chronological information between sequential posts, and 
\item prove it to be beneficial in almost all examined architectures.
\end{enumerate*}

\section*{Acknowledgements}
We would like to thank the anonymous reviewers
for their constructive feedback. \revthree{Moreover, we would like to thank Christos Xypolopoulos and Giannis Nikolentzos for providing 
\begin{enumerate*}[label=(\roman*)]
\item the Twitter dataset (tweet ids) and 
\item instructions to reproduce the results of their graph-based approach.
\end{enumerate*}}

\bibliography{naaclhlt2019}
\bibliographystyle{acl_natbib}




\end{document}